\documentclass{svproc}
\usepackage{times}

\usepackage{soul}
\usepackage{url}
\usepackage[hidelinks]{hyperref}
\usepackage[utf8]{inputenc}
\usepackage[small]{caption}
\usepackage{graphicx}
\usepackage{amsmath}
\usepackage{booktabs}

\usepackage{amsmath,amssymb,amsfonts}
\usepackage{algorithmic}
\usepackage{graphicx}
\usepackage{subfiles}
\usepackage{textcomp}
\usepackage{xcolor}
\usepackage{caption} 
\usepackage[misc]{ifsym}

\usepackage{url}

\makeatletter
\newcommand{\printfnsymbol}[1]{%
  \textsuperscript{\@fnsymbol{#1}}%
}
\makeatother

\begin{document}
\mainmatter              
\title{Logistics, Graphs, and Transformers: \\
Towards improving Travel Time Estimation
}
\titlerunning{Logistics, Graphs, and Transformers:
Towards improving Travel Time Estimation}  
%

\author{Natalia Semenova\inst{1,2}(\Letter) \and
Vadim Porvatov\inst{1,3}\and
Vladislav Tishin\inst{1,3}\and \\
Artyom Sosedka\inst{1,3}\and 
Vladislav Zamkovoy\inst{1}
}
\authorrunning{N. Semenova et al.} 
%
%
\institute{Sberbank, Moscow 117997, Russia, \and
Artificial Intelligence Research Institute, Moscow 105064, Russia, \and
National University of Science and Technology ``MISIS'', Moscow 119991, Russia
\email{semenova.bnl@gmail.com}
\url{}}

\maketitle          

\begin{abstract}

The problem of travel time estimation is widely considered as the fundamental challenge of modern logistics. The complex nature of interconnections between spatial aspects of roads and temporal dynamics of ground transport still preserves an area to experiment with. However, the total volume of currently accumulated data encourages the construction of the learning models which have the perspective to significantly outperform earlier solutions. In order to address the problems of travel time estimation, we propose a new method based on transformer architecture -- TransTTE. 

\end{abstract}
\section{Introduction}

As long as ground transport dramatically increases its quantitative presence in the cities, traffic management becomes more complex and hence less predictable for drivers. In order to handle the escalation of such a negative trend, it is important to effectively estimate the essential parameters describing traffic dynamics. One of the most important values among such quantities is the estimated time of arrival (ETA) which could be considered as the expected time expenditure for a trip between two locations. Accurate travel time estimation (TTE) is mostly challenging for cars due to the presence of the extensive limitations induced by the road network structure. These aspects of urban traffic ordinary require special spatio-temporal methods implementation to be handled.
The contributions of our work are the following:

\begin{enumerate}
    \item We proposed TransTTE model that could utilize the spatio-temporal dependencies and explored the capabilities of the transformer model application in the domain of TTE via comparison with several baselines. Computational experiments allow us to conclude that the TransTTE architecture achieves competitive performance. 
    \item We published a new dataset related to the city of Omsk.  
    \item For the needs of demonstration, we developed a web service based on the TransTTE model.
\end{enumerate}

The application is available at \url{http://transtte.online} and the code could be accessed from \url{https://github.com/Vloods/TransTTE_demo}.

\section{Related Work}
Generally, the TTE methods could be divided into two categories related to the different approaches of ETA computing. The first one is based on the extraction of the total traveled time regarding each segment of the path~\cite{example:segment_1}. Such models do not capture any global properties of the path which is explicitly reflected in their performance. The second class of methods utilizes the corresponding trip path as a whole~\cite{STGNN-TTE}. The results achieved by these approaches defined the mainstream of current research in the domain of TTE and hence are widely established in the experiments section.

\section{Framework design}
In this section, we discuss the aspects of model design and deployment as a part of a developed web service.  

\textbf{Task}. Given an origin, destination, and departure time, our goal is to estimate the duration using the set of historical trip dataset \textit{X} and the underlying road network \textit{G}.


\textbf{Model}. The transformer architecture has recently become a prevalent approach in many domains, such as natural language processing and computer vision. Yet, it has not achieved competitive performance on popular leaderboards of graph-level prediction compared to mainstream GNN variants. Therefore, it remains a question of how transformers could perform well for graph representation learning in the TTE task. 

One of the graph-oriented aspects in the Graphormer architecture~\cite{graphormer} is a \emph{centrality encoding} which assigns each node two real-valued embedding vectors according to its indegree and outdegree $h_{i}^{(0)}=x_{i}+z^{-}+z^{+}$, where $z^{-}, z^{+} \in \mathbb{R}^d$ are learnable embedding vectors specified by the indegree $\text{deg}^{-}(v_i)$ and outdegree $\text{deg}^{+}(v_i)$ respectively. Along with centrality encoding, \emph{spatial encoding} is used to capture the structural relation via function $\phi(v_i, v_j):V\times V\to\mathbb{R}$ which measures the spatial relation between nodes $v_i$ and $v_j$ of road network $G$. Original choice of such function $\phi(v_i,v_j)$ is the shortest distance between $v_i$ and $v_j$ which further serves as a bias term in the self-attention module $A_{ij}=\frac{(h_iW_{Q})(h_jW_{K})^T}{\sqrt{d}}  + b_{\phi(v_i,v_j)}$, where $b_{\phi(v_i,v_j)}$ is a learnable scalar indexed by $\phi(v_i,v_j)$ which is shared across all layers.

\textbf{Data}. In this paper, we use the dataset related to the road networks of Abakan and Omsk\footnote{The full data could be requested from semenova.bnl@gmail.com} which have different scales and road topology~\cite{Hybrid}. The datasets are collected in a monthly period starting from December 1, 2020. The datasets consist of a road network with corresponding to its segments features and trip part. The presence of the noisy data among the trip part of the datasets encourages us to apply filtering regarding the rebuild count feature, maximum/minimum length, and the total time of the trips. 

\begin{figure}[t]
\includegraphics[width=\textwidth]{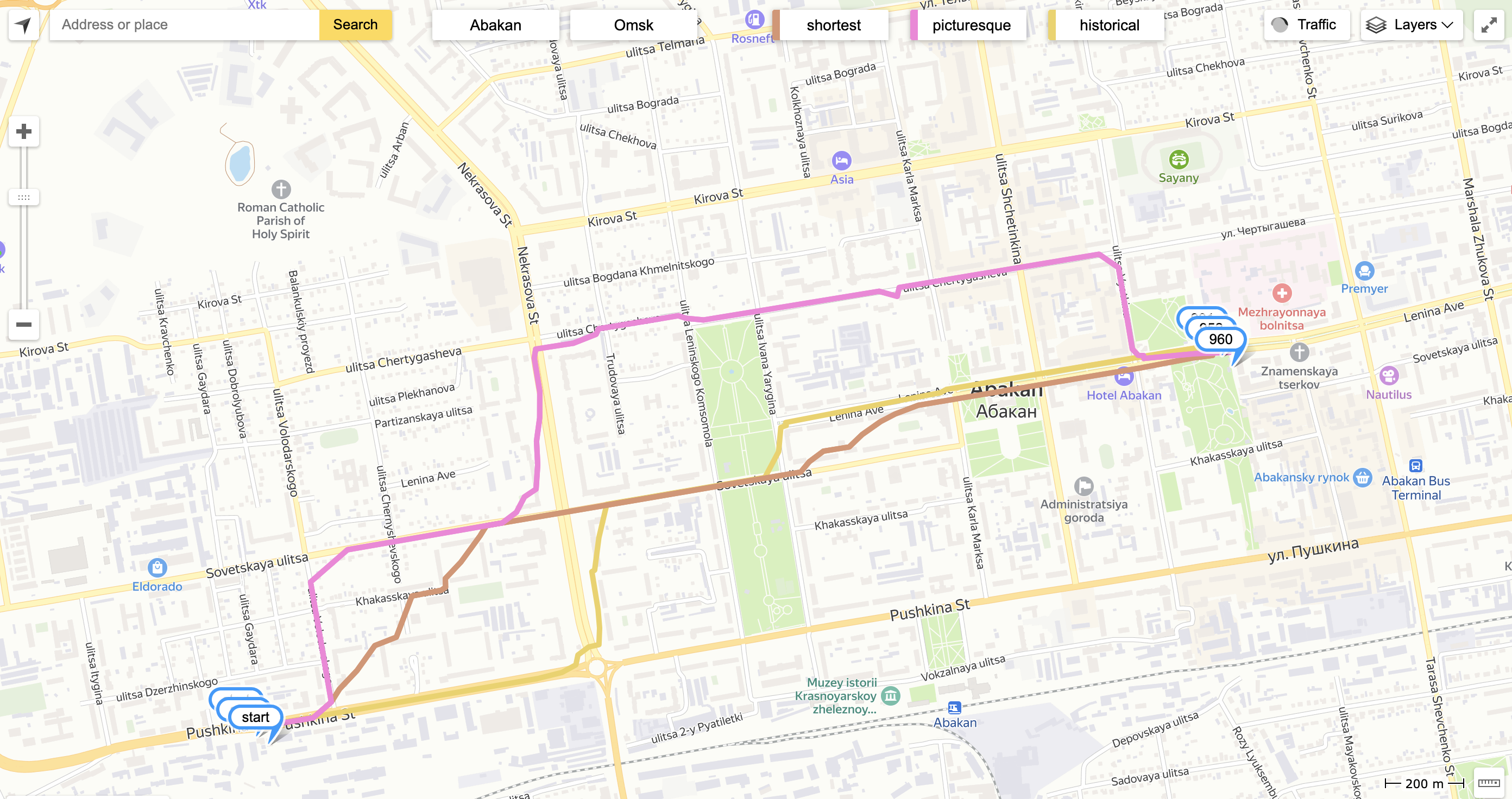}
\caption{Application interface.} 
\label{interface}
\end{figure}

\textbf{Web application}. For the needs of model demonstration, we deployed our TTE service in the platform of the Yandex Maps project, Figure~\ref{interface}. The interface allows choosing between two cities and three types of routes regarding demands of a user.

\section{Results}
In addition to the computing TTE for the shortest route, the framework also evaluates the routes based on the metrics of picturesqueness and historicity. Using the OpenStreetMap API, we managed to parse information about the location of historical, cultural and natural objects. The number of certain objects is used as the road segments weights for Dijkstra's algorithm $W_i = \frac{1}{1 + C_r}$, where $W_i$ is the weight for the $i$'th segment of the road, $C_r$ is the number of objects within a radius $r$ of a road segment.

We reimplemented Graphormer architecture to accelerate the training process and consider the peculiar properties of road trips. Due to caching spatial encoding values, we were able to speed up training by almost 10 times. Several baselines for TTE task were also implemented to verify the effectiveness of the proposed model, Table~\ref{tab:results}. We considered results made by gradient boosted decision trees along with the more sophisticated pipelines. WDR~\cite{wdr} uses generalized linear model and LSTM together to compute travel time. MURAT~\cite{murat}, in its turn, produces unsupervised representations due to DeepWalk~\cite{deepwalk} and applies residual feedforward blocks to predict travel time and distance. 

The best result was achieved on Graphormer$_{\small \textsc{Slim}}$ ($L=12, d=80$) version with reduced size of dimension.
As the optimizer we used AdamW~\cite{AdamW}. Experiments were done with 5 Tesla V100 GPUs and 460 Gb of RAM. The training time of the different  configurations of TransTTE lies in the interval from 2.5 hours up to 5 hours which is smaller than in case of WDR (7 hours) and MURAT (5.5 hours).

\section{Conclusion and Outlook}

\begin{table}[t] 
\centering
\begin{tabular}{c|c|cc|cc|cc|cc}
\toprule
\toprule
\multicolumn{2}{c|}{ Dataset } & \multicolumn{4}{c|}{Omsk} & \multicolumn{4}{c}{Abakan} \\
\midrule
\multicolumn{2}{c|}{Split} & \multicolumn{2}{c|}{Train} & \multicolumn{2}{c|}{Test} & \multicolumn{2}{c|}{Train} & \multicolumn{2}{c}{Test}\\
\midrule
\multicolumn{2}{c|}{} & MAE & RMSE & MAE & RMSE & MAE & RMSE & MAE & RMSE\\
\midrule
\multicolumn{2}{c|}{GBDT} & 403.921 & 582.011 & 408.644 & 573.559 & 244.119 & 449.250 & 248.862 & 399.534\\
\multicolumn{2}{c|}{MURAT} & 279.616 & 438.228 & 286.491 & 443.397 & 179.037 & 285.003 & 185.153 & 286.934 \\
\multicolumn{2}{c|}{WDR} & 311.581 & 440.511 & 336.756 & 487.876 & 173.684 & 285.132 & 182.296 & 293.551 \\
\midrule
\multicolumn{2}{c|}{TransTTE} & 101.381 & 387.241 & 105.464 & 261.103 & 81.048 & 285.032 & 83.616 & 168.421\\
\bottomrule
\bottomrule
\end{tabular}
\caption{Evaluation of different pipelines and comparison with proposed method}
\label{tab:results}
\end{table}

In this paper, we proposed the new transformer-based approach to the computing of ETA and explored its performance. The experiments revealed the perspective of graph transformer utilization in the travel time estimation.
In the upcoming studies, we want to extend the current transformer architecture by virtue of extra road network features and more precise work with the temporal aspect of road trips. Future research should be devoted to the development of a joint TransTEE model which could compute travel time for rides on different city networks indeed.
%
%

\bibliographystyle{splncs03} 
\bibliography{8-refs}

\begin{thebibliography}{1}
\providecommand{\url}[1]{\texttt{#1}}
\providecommand{\urlprefix}{URL }

\bibitem{example:segment_1}
Asghari, M., Emrich, T., Demiryurek, U., Shahabi, C.: Probabilistic estimation
  of link travel times in dynamic road networks. pp. 1--10 (11 2015)

\bibitem{STGNN-TTE}
Jin, G., Wang, M., Zhang, J., Sha, H., Huang, J.: Stgnn-tte: Travel time
  estimation via spatial–temporal graph neural network. Future Generation
  Computer Systems  126,  70--81 (2022),
  \url{https://www.sciencedirect.com/science/article/pii/S0167739X21002740}

\bibitem{murat}
Li, Y., Fu, K., Wang, Z., Shahabi, C., Ye, J., Liu, Y.: Multi-task
  representation learning for travel time estimation. In: International
  Conference on Knowledge Discovery and Data Mining (KDD '18) (2018)

\bibitem{AdamW}
Loshchilov, I., Hutter, F.: Decoupled weight decay regularization (2019)

\bibitem{deepwalk}
Perozzi, B., Al-Rfou, R., Skiena, S.: Deepwalk: Online learning of social
  representations. In: Proceedings of the 20th ACM SIGKDD international
  conference on Knowledge discovery and data mining. pp. 701--710 (2014)

\bibitem{Hybrid}
Porvatov, V., Semenova, N., Chertok, A.: Hybrid graph embedding techniques
  in estimated time of arrival task. In: Benito, R.M., Cherifi, C., Cherifi,
  H., Moro, E., Rocha, L.M., Sales-Pardo, M. (eds.) Complex Networks {\&} Their
  Applications X. pp. 575--586. Springer International Publishing, Cham (2022)

\bibitem{wdr}
Wang, Z., Fu, K., Ye, J.: Learning to estimate the travel time. In: Proceedings
  of the 24th ACM SIGKDD International Conference on Knowledge Discovery and
  Data Mining. p. 858–866. KDD '18, Association for Computing Machinery, New
  York, NY, USA (2018), \url{https://doi.org/10.1145/3219819.3219900}

\bibitem{graphormer}
Ying, C., Cai, T., Luo, S., Zheng, S., Ke, G., He, D., Shen, Y., Liu, T.Y.: Do
  transformers really perform bad for graph representation? arXiv preprint
  arXiv:2106.05234  (2021)

\end{thebibliography}

\end{document}